\documentclass[journal, letterpaper]{IEEEtran_old}

\usepackage{packages}
\usepackage{textcomp}

\def\BibTeX{{\rm B\kern-.05em{\sc i\kern-.025em b}\kern-.08em
    T\kern-.1667em\lower.7ex\hbox{E}\kern-.125emX}}

\newtheorem{theorem}{Theorem}

\newtheorem{lemma}{Lemma}
\newtheorem{assumption}{Assumption}
\newtheorem{remark}{Remark}

\usepackage{tikz}
\newcommand*\circled[2]{\tikz[baseline=(char.base)]{
\node[circle,draw,scale=#2,inner sep=1.2pt] (char) {#1};}}

\newcommand{\norm}[1]{\left\|#1\right\|}

\renewcommand\labelenumi{(\roman{enumi})}
\renewcommand\theenumi\labelenumi

\algnewcommand{\Initialize}[1]{%
  \State \textbf{Initialization:}
  \Statex \hspace*{\algorithmicindent}\parbox[t]{.8\linewidth}{\raggedright #1}
}
\algnewcommand{\Input}[1]{%
  \State \textbf{Input:}
  \Statex \hspace*{\algorithmicindent}\parbox[t]{.8\linewidth}{\raggedright #1}
}

\begin{document}

\title{FedZeN: Towards superlinear zeroth-order federated learning via incremental Hessian estimation}

% Alessio Maritan, \IEEEmembership{Graduate Student Member, IEEE}
% Luca Schenato, \IEEEmembership{Fellow, IEEE}
% Subhrakanti Dey, \IEEEmembership{Senior Member, IEEE}

\author{Alessio Maritan$^1$, Subhrakanti Dey$^2$, and Luca Schenato$^1$%
\thanks{$^1$ A. Maritan and L. Schenato are with the Department of Information Engineering, University of Padova, Italy. Email: alessio.maritan@phd.unipd.it, schenato@dei.unipd.it.}
\thanks{$^2$ Subhrakanti Dey is with the Department of Electrical Engineeringe, Uppsala University, Sweden. Email: subhrakanti.dey@angstrom.uu.se}}

%\author{\IEEEauthorblockN{Alessio Maritan$^*$, Luca Schenato$^*$, and Subhrakanti Dey$^\dag$}
%\IEEEauthorblockA{$^*$Department of Information Engineering, University of Padova, Italy \\
%$^\dag$Department of Electrical Engineeringe, Uppsala University, Sweden \\
%Email: alessio.maritan@phd.unipd.it, schenato@dei.unipd.it, subhrakanti.dey@angstrom.uu.se}
%}
\maketitle
\thispagestyle{empty}

%+++++++++++++++++++++++++++++

\begin{abstract}
Federated learning is a distributed learning framework that allows a set of clients to collaboratively train a model under the orchestration of a central server, without sharing raw data samples.
Although in many practical scenarios the derivatives of the objective function are not available, only few works have considered the federated zeroth-order setting, in which functions can only be accessed through a budgeted number of point evaluations.
In this work we focus on convex optimization and design the first federated zeroth-order algorithm to estimate the curvature of the global objective, with the purpose of achieving superlinear convergence. 
We take an incremental Hessian estimator whose error norm converges linearly, and we adapt it to the federated zeroth-order setting, sampling the random search directions from the Stiefel manifold for improved performance.
In particular, both the gradient and Hessian estimators are built at the central server in a communication-efficient and privacy-preserving way by leveraging synchronized pseudo-random number generators.
We provide a theoretical analysis of our algorithm, named FedZeN, proving local quadratic convergence with high probability and global linear convergence up to zeroth-order precision.
Numerical simulations confirm the superlinear convergence rate and show that our algorithm outperforms the federated zeroth-order methods available in the literature.
\end{abstract}


\section{Introduction}
Federated learning (FL) is a large-scale learning framework that allows multiple users to collaboratively train machine learning models while preserving the individual privacy.
The goal is to expose the model to as much data as possible, achieving better generalization capabilities than if each client trains a separate model on his own data.
Clients never transmit their raw data samples over the network, but rather exchange model updates with a central orchestrating server. This can dramatically reduce the communication cost of the learning process and provides some degrees of data security, which can be further improved by incorporating mechanisms such as differential privacy and homomorphic encryption. Moreover, the distributed nature of FL allows to overcome the limited scalability of the standard centralized setting, in which all the training data must be gathered and processed at a single machine with enough computational power and storage resources. For these reasons FL is the tool of choice when the training data is naturally distributed in form of data islands, which often happens in networks of smartphones, IoT sensors or other devices.

In many relevant cases, such as simulation-based or black-box optimization, the derivatives of the objective functions may be expensive or infeasible to obtain \cite{book_DFBO}. Most of the existing federated learning algorithms, including the well-known FedAvg \cite{McMahan_Moore_Ramage_Hampson_Arcas_2017}, are gradient-based and thus cannot be applied in such situations. A possible solution is offered by the class of zeroth-order (ZO) algorithms, that do not require any knowledge of the function derivatives. Rather, they only need the objective to be evaluated at certain query points, and they estimate derivatives by mean of finite-differences along a set of search directions.
We address the reader to \cite{liu2020primer} for a survey on general zeroth-order optimization, and below we briefly review the ZO federated algorithms available in the literature: FedZO \cite{Fang_Yu_Jiang_Shi_Jones_Zhou_2022} is a zeroth-order version of FedAvg; ZONE-S \cite{Hajinezhad_Hong_Garcia_2019} is a primal-dual algorithm in which at each iteration only one client is active, and the central server minimizes an augmented Lagrangian function; BAFFLE \cite{Feng_Pang_Du_Chen_Yan_Lin_2023} uses a stochastic gradient estimator based on Stein's identity and focuses on the privacy aspect; AsyREVEL \cite{Zhang_Gu_Dang_Deng_Huang_2021} addresses the vertical FL scenario, while this work concerns the horizontal FL setting.

Remarkably, none of the above algorithms considers the curvature of the objective function, missing out on the possibility to greatly improve the convergence rate. In fact, preconditioning with the Hessian matrix often leads to larger improvements per iteration and consequently much fewer iterations needed to converge. This is especially desirable in FL, where many communication rounds are generally needed, and could sensibly reduce bandwidth consumption and idle time. ZO-JADE \cite{Maritan_Schenato_2023}, which is the only distributed zeroth-order algorithm to exploit the curvature information, estimates both the gradient and the diagonal of the Hessian matrix computing central-differences along the canonical basis. However, neglecting the off-diagonal elements of the Hessian may lead to suboptimal performance when the objective function is highly skewed. Moreover, ZO-JADE is designed for a general mesh network and does not take full advantage of the star topology of the federated setting.

Looking outside the zeroth-order literature, there are two second-order federated algorithms that provide superlinear convergence, namely FedNL \cite{Safaryan_Islamov_Qian_Richtárik_2022} and SHED \cite{Fabbro_Dey_Rossi_Schenato_2022}. However both do not support approximate derivatives, preventing straightforward zeroth-order implementations where the exact gradient and Hessian are replaced with estimates.

\textbf{Contributions}: Motivated by the absence of federated zeroth-order algorithms which leverage the curvature information, in this paper we design a novel algorithm, named FedZeN (Zeroth-order Newton).
We focus on convex optimization problems and aim to achieve superlinear convergence, which requires knowledge of the full Hessian matrix. For this reason, we extend the randomized incremental estimator proposed in \cite{Leventhal_Lewis_2011} to make it suitable for federated zeroth-order implementation.
In particular, we exploit synchronized pseudo-random number generators to sample a common set of $r$ search direction at all the nodes. The clients query their local functions according to the search directions, compute a set of coefficients needed to build both the gradient and Hessian estimators, and send them to the central server. The latter updates the model parameters using a Newton-type method, which is known to be significantly faster than first-order methods. The approximation error due to the zeroth-order estimation is handled using either appropriate regularization or an eigenvalue clipping safeguarding.

Below we list the main novelties and distinguishing features of the proposed method.
\textit{(i)} We devise a federated incremental estimator of the full Hessian matrix, which enables tackling federated optimization problems using second-order methods even when the exact derivatives are not available. Our estimator is the distributed zeroth-order counterpart of the one proposed in \cite{Leventhal_Lewis_2011}, that converges almost surely to the true Hessian and whose error norm goes to zero linearly in expectation. We propose to generate the search directions needed by the estimator by uniformly sampling the Stiefel manifold, which empirically provides better accuracy and enables the use of an excellent gradient estimator.
\textit{(ii)} We design FedZeN, the first federated zeroth-order algorithm to estimate and exploit the Hessian of the global objective function. We provide a theoretical analysis of the algorithm, proving local quadratic convergence with high probability and global linear convergence up to zeroth-order precision. Our numerical simulations show that FedZeN outperforms the existing federated zeroth-order algorithms and exhibits superlinear convergence.
\textit{(iii)} The proposed distributed derivative estimation procedure naturally addresses some important concerns in federated learning. The first is non-identically distributed data: the algorithm can be applied to pools of clients with heterogeneous data distributions and is unaffected by client drift. The second is privacy: if the internal seed of the pseudo-random generators is kept private, the proposed procedure hides the estimated derivatives from potential external eavesdroppers. 
Regarding the computational and communication costs, at each iteration clients only need to evaluate their local function at $2r+1$ query points and transmit to the central server $d+r$ scalar values, where $d$ is the dimension of the problem. The design parameter $r$ is independent from the dimension of the problem, making the algorithm suitable for client devices with limited resources.

\textbf{Notation}: We denote with $I_d$ the $d$-dimensional identity matrix and with $\mathbb{E}[\cdot]$ the expectation. Given a matrix, $\left\| \cdot \right\|$  is the spectral norm while $\left\| \cdot \right\|_F$ is the Frobenius norm. For brevity, we indicate with $[n]$ the set of integers $\{1, \dots, n\}$, and with $\mathcal{U}(\mathbb{S})$ the uniform distribution on the unit sphere $\mathbb{S} = \{ z \in \mathbb{R}^d \text{ s.t. } \norm{z} = 1 \}$.

\section{Problem Formulation}
We consider the horizontal federated learning setting, where local datasets consist of samples with different IDs that belong to the same feature space. Data is not independently or identically distributed, i.e. the data distribution can vary across clients.
We consider a federation of $n$ clients wanting to collaboratively train a model parametrized by $x \in \mathbb{R}^d$ by solving the empirical risk minimization
\begin{equation}
    f(x^\star) = \min_{x \in \mathbb{R}^d} \left\{ f(x) \coloneqq \frac{1}{n} \sum_{i=1}^n f_i(x) \right\}.
    \label{eq:global_function}
\end{equation}
Here $f_{i}(x): \mathbb{R}^d \rightarrow \mathbb{R}$ is the loss function of client $i$, and the global average $f(x)$ satisfies the following assumption, which is standard in convex optimization.

\begin{assumption}
    The global cost is $m$-strongly convex and twice continuously differentiable with Lipschitz derivatives, i.e. there exist positive constants $m, L_0, L_1, L_2$ such that $\forall x,y \in \mathbb{R}^d$
    \begin{equation*}
        \begin{aligned}
            \norm{f(x) - f(y)} & \leq L_0 \norm{x-y}, \\
            \norm{\nabla^2 f(x) - \nabla^2 f(y)} & \leq L_2 \norm{x-y}, \\
            m I_d \leq \nabla^2 f(x) & \leq L_1 I_d.
        \end{aligned}
    \end{equation*}
    \label{assumption:Lipschitz}
\end{assumption}

In order to apply our distributed derivative estimation technique, we assume that all the clients and the central server own the same deterministic pseudo-random number generator (PRNG). This trick allows to generate common vectors at all the devices by just periodically sending seeds or internal states to ensure synchronization, greatly reducing the communication overhead of the algorithm. The presence of PRNGs is a mild requirement and is assumed also in other works, such as \cite{Feng_Pang_Du_Chen_Yan_Lin_2023} and \cite{Agafonov_Kamzolov_Tappenden_Gasnikov_Takáč_2022}.

\begin{assumption}[Synchronized PRNG]
    All the clients and the central server are equipped with the same pseudo-random number generator, whose output sequence can be determined a priori by having knowledge of the internal seed.
    \label{assumption:PRNG}
\end{assumption}

\section{Zeroth-order Oracles}
Zeroth-order estimators approximate derivatives by means of finite-differences between values taken by the objective function at given query points. The latter are chosen in a neighborhood of the current model parameters, fixing a set of search directions and a small scalar $\mu > 0$. The value of the finite-difference granularity $\mu$, also called discretization or smoothing parameter, is usually chosen based on the specific application and on the machine precision.
When choosing the derivative estimator, one must consider the level of accuracy but also the associated computational cost. In fact, the common assumption in the ZO optimization field is that not only the exact derivative is inaccessible or prohibitive to obtain, but also that function evaluations are expensive and possibly budgeted in number. 

% +++++++++++++++++++

\subsection{Incremental randomized Hessian estimator}

Most ZO algorithms avoid estimating the Hessian matrix, as this typically requires much more function evaluations than gradient estimation. For example, to approximate all the entries of the Hessian using forward finite-differences along the canonical basis $\{e_1, \dots, e_d \}$, one has to query the objective function at the points $x$, $\{ x + \mu e_i \}$, $\{ x + \mu e_i + \mu e_j\}$ with $i,j \in [d]$, for a total of $(d+1)(d/2 + 1)$ evaluations \cite{nocedal1999numerical}. If instead the distinct off-diagonal entries are estimated through symmetric differences between the points $\{ x \circ \mu e_i \diamond \mu e_j\}$ for $i,j \in [d]$, for $\circ, \diamond \in \{+,-\}$, then the number of function evaluations grows to $2d^2 + 1$. 
% 4*(n^2 - n)/2 (off-diagonal) + 2n+1 (diagonal)
% https://www.sfu.ca/sasdoc/sashtml/iml/chap11/sect8.htm

To avoid necessarily computing $O(d^2)$ function values, one can resort to randomized estimation schemes, that allow to choose an arbitrary number $r$ of search directions at the cost of a possibly larger approximation error.
An example of randomized Hessian estimator is the one proposed in \cite{Feng_Wang_2023}, which performs $4r^2$ function queries along orthogonal directions sampled from the Stiefel manifold. The error norm of this estimator is shown to decrease sublinearly with the number of search directions, and to suddenly drop only when $r=d$. % see Figure 6
Based on the second-order Stein's identity, \cite{Balasubramanian_Ghadimi_2022} develops some unbiased estimators of the Hessian of a Gaussian-smoothed version of the objective function. However, as shown in our simulations, in practice these estimators still require too many function evaluations to provide an acceptable estimate. In fact, being sample averages over the set of search directions, by the law of large numbers their variance decreases with sublinear rate $1/r$.

In this work we employ an Hessian estimator based on a different principle. Given an initial symmetric matrix $H^0 \in \mathbb{R}^{d \times d}$, we apply $r$ times the update proposed in \cite{Leventhal_Lewis_2011}
\begin{equation}
    H^k = H^{k-1} + (u^T \nabla^2 f(x) u - u^T H^{k-1} u) u u^T ,
    \label{eq:Leventhal}
\end{equation}
where $u \sim \mathcal{U} (\mathbb{S})$.
The idea behind this iterative formula is to add a rank-one matrix such that the updated estimator matches the true Hessian along the direction $u$. The recursion \eqref{eq:Leventhal} satisfies the linear convergence condition
\begin{equation}           
    \mathbb{E} \left[ \norm{ H^k - \nabla^2 f(x)}_F^2 \right]
    \leq \eta \norm{ H^{k-1} - \nabla^2 f(x) }_F^2,
    \label{eq:Leventhal_convergence_rate}
\end{equation}
where $\eta = 1 - 2/(d^2+2d)$, and asymptotically $H^k$ converges almost surely to the exact Hessian \cite{Leventhal_Lewis_2011}.
Since the true Hessian is obviously not available, as also mentioned in \cite{Leventhal_Lewis_2011} Hessian-vector products can be estimated using finite-differences, which makes the update \eqref{eq:Leventhal} ideal for zeroth-order optimization.
In particular, in FedZeN we approximate the directional curvature as
\begin{equation*}
    u^T \nabla^2 f(x) u
    \approx \frac{f(x + \mu u)-2f(x)+f(x - \mu u)}{\mu^2} ,
\end{equation*}
and therefore computing the Hessian estimator $H^r$ requires $2r+1$ function evaluations.

Differently from the other estimators available in the literature, \eqref{eq:Leventhal} is an incremental formula. This lends itself to warm-start the estimator by initializing it with the estimate from the previous iteration. This is especially useful when the Hessian is constant or slowly changing and when approaching the global optimum, so that only few updates per iteration are needed. On the contrary, the estimators in \cite{Feng_Wang_2023} and \cite{Balasubramanian_Ghadimi_2022} are designed to be reset at each iteration and not to exploit past estimates, and in this way they lose all previously collected information.

% ++++++++++++++++++++

\subsection{Stiefel sampling}
The first and fundamental step to build the randomized Hessian estimator is to choose a set of search directions $ \{u_j \sim \mathcal{U} (\mathbb{S}) \}$, $j \in [r]$. The standard way to generate these directions is to sample $r$ vectors from $\mathcal{N}(0, I_d)$ and project them on the unit hypersphere by dividing by their norm. However, this sampling procedure is only asymptotically optimal, and for limited values of $r$ may cause to oversample some regions of the space while barely exploring others. To address this problem, in FedZeN we generate a matrix uniformly sampled from the Stiefel manifold
\begin{equation*}
    V_{r,d} = \{ U \in \mathbb{R}^{d \times r} \text{ such that } U^T U = I_d \}
\end{equation*}
and use its $r \leq d$ columns as search directions. Intuitively, since this set of vectors is orthogonal it should be more evenly spread in the search space, thus maximizing the information gain and reducing redundancy. Most importantly, the marginal distribution of these vectors is $\mathcal{U} (\mathbb{S})$, which is the one required by our Hessian estimator. An explanation of why this last fact holds is provided in \cite{Feng_Wang_2023}, which first introduced Stiefel sampling for zeroth-order optimization. The procedure to uniformly sample from the Stiefel manifold is based on Theorem 2.2.1 of \cite{chikuse2003statistics}, stating that a matrix $U = [u_1, \dots, u_r]$ uniformly distributed on $V_{r,d}$ can be expressed as 
\begin{equation}
    U = X (X^T X)^{-1/2},
    \;
    X \in \mathbb{R}^{d \times r} \text{ s.t. } X_{ij} \overset{i.i.d.}{\sim} \mathcal{N}(0, 1) .
    \label{eq:Stiefel}
\end{equation}
In our tests, generating the search directions according to \eqref{eq:Stiefel} instead of using non-orthogonal directions considerably improves the accuracy of the Hessian estimator, especially for small values of $r$. In case $r > d$, we generate $\lceil r/d \rceil$ separate orthogonal matrices.

% ++++++++++++++++++++

\subsection{Zeroth-order gradient estimator}

In this work we adopt the gradient estimator
\begin{equation}
    g(x) = \sum_{j=1}^d \frac{ f(x + \mu u_j) - f(x - \mu u_j)}{2 \mu} u_j
    \label{eq:gradient_estimator}
\end{equation}
where the orthonormal search directions $\{ u_j \sim \mathcal{U} (\mathbb{S}) \}$, $j \in [d]$ are a subset of the directions used to build the Hessian estimator.
The choice of this gradient estimator is motivated by several reasons. First, as constructing the Hessian estimator $H^r$ involves querying the objective function at the points $\{ x \pm \mu u_j \}$, $j \in [r]$, it makes sense to reuse these function values to estimate also the gradient for free.
Second, numerical simulations show that in practice $r$ must be at least greater than the dimension of the problem to get a satisfactory approximation of the Hessian, which guarantees that $d$ orthonormal search directions are always available.
Finally, by estimating the gradient along an orthonormal basis we can provide deterministic guarantees on the approximation error, as shown by the following Lemma.
\begin{lemma}[Error of the gradient estimator]
    If the set of search directions used to build the gradient estimator \eqref{eq:gradient_estimator} is an orthonormal basis $U = \{u_1, \dots, u_d\}$, then $\forall x \in \mathbb{R}^d$
    \begin{equation*}
            \norm{\nabla f(x) - g(x)}
            \leq \frac{d L_2 \mu^2}{6}.
    \end{equation*}
    \label{lemma:gradient_estimator_error}
\end{lemma}

\begin{proof}
    Define $\Delta_i = \nabla^2 f(x + t\mu u_i) - \nabla^2 f(x - t\mu u_i)$. Since $U$ is a basis of $\mathbb{R}^d$ and $\norm{u_i}=1$ $\forall i \in [d]$, using Taylor expansion with integral remainder we get
    \begin{equation*}
        \begin{aligned}
            & \norm{\nabla f(x) - g(x)}
            = \norm{ \sum_{i=1}^d (\nabla f(x)^T u_i) u_i - g(x)} \\
            & \leq \sum_{i=1}^d \norm{ \left( \nabla f(x)^T u_i - \frac{ f(x + \mu u_i) - f(x - \mu u_i)}{2 \mu} \right) u_i} \\
            &= \sum_{i=1}^d \left| \frac{\mu}{2} u_i^T \int_0^1 (1-t) \Delta_i dt \; u_i \right| \norm{u_i} \\
            &\leq \sum_{i=1}^d \left| \frac{\mu}{2} \norm{u_i}^2 \int_0^1 (1-t) L_2 \norm{2 t\mu u_i} dt \right|
            = \frac{d \mu^2 L_2}{6}.
         \end{aligned}
    \end{equation*}
\end{proof}

Other properties of the estimator \eqref{eq:gradient_estimator} can be found in \cite{Feng_Wang_2023}. In comparison, commonly used randomized gradient estimators such as the ones employed in FedZO \cite{Fang_Yu_Jiang_Shi_Jones_Zhou_2022} and ZONE-S \cite{Hajinezhad_Hong_Garcia_2019} do not search along orthogonal directions and are associated with larger variance and approximation errors \cite{Feng_Wang_2023} \cite{berahas2022theoretical}.

\section{Federated Hessian estimation}
In this section we describe how to build the randomized estimators in a communication-efficient way by taking advantage of the star topology of the network, and we introduce the proposed federated zeroth-order algorithm.
We use the subscript $k$ where needed to denote the value of a variable at the $k$-th iteration of the algorithm.

According to Assumption \ref{assumption:PRNG}, each client can access a pseudo-random number generator, and all the generators can be synchronized by making the central server broadcast a common internal seed at the first iteration. In the initialization step of the algorithm the master also chooses the initial Hessian estimator, which can be any symmetric matrix $H_1^0 \in \mathbb{R}^{d \times d}$, for example $H_1^0 = \beta I_d$ with $\beta>0$.
At each iteration, both the clients and the master use their PRNG to generate a common random matrix $X \in \mathbb{R}^{d \times r}$ such that $X_{ij} \overset{i.i.d.}{\sim} \mathcal{N}(0, 1)$. From the latter, using Stiefel sampling \eqref{eq:Stiefel} they compute the set of vectors $\{ u_j \sim \mathcal{U} (\mathbb{S}) \}$, $j \in [r]$, which is the same at all the nodes. The central server broadcasts the current decision vector $x_k$. Each client $i$ evaluates his local function at the points $x_k$ and $\{ x_k \pm \mu u_j \}$, $j \in [r]$ to compute the $d$ gradient coefficients
\begin{equation}
    c_{ij} = \frac{ f_i(x_k + \mu u_j) - f_i(x_k - \mu u_j)}{2 \mu} ,
    \label{eq:c_ij}
\end{equation}
and the $r$ directional curvatures
\begin{equation}
    b_{ij} = \frac{f_i(x_k + \mu u_j)-2f_i(x_k)+f_i(x_k - \mu u_j)}{\mu^2} .
    \label{eq:b_ij}
\end{equation}
These $d+r$ scalars are sent to the master, where they are averaged over the set of clients.
Using the fact that the search directions are the same for all nodes, the central server is able to build the derivative estimators, where the Hessian estimator is updated starting from $H_k^0 = H_{k-1}^r$ when $k>1$.
\begin{equation}
    g_k = \sum_{j=1}^d \left( \frac{1}{n} \sum_{i=1}^n c_{ij} \right) u_j ,
    \label{eq:gradient_estimator_federated}
\end{equation}
\begin{equation}
    H_k^j = H_k^{j-1} + \left( \frac{1}{n} \sum_{i=1}^n b_{ij} - u_j^T H_k^{j-1} u_j \right) u_j u_j^T ,
    \;
    j \in [r].
    \label{eq:Leventhal_federated}
\end{equation}

To get the new model parameters, the central server performs a step of a Newton-type method, which requires $H_k^r$ to be invertible and positive-definite. To ensure that this constraint is satisfied and improve the robustness of the algorithm with respect to estimation errors, we consider two possible safeguarding mechanisms. The first is a simple and computationally inexpensive regularization, where a scalar multiple of the identity matrix is added to $H_k^r$: 
\begin{equation}
    Z_k = (H_k^{r} + \rho I_d)^{-1},
    \quad \rho > 0 .
    \label{eq:H_rhoIdentity}
\end{equation}
% https://en.wikipedia.org/wiki/Eigendecomposition_of_a_matrix#Real_symmetric_matrices
The second is eigenvalue clipping, based on the spectral decomposition $H_k^r = QDQ^T$, where $D$ is the diagonal matrix whose entries are the eigenvalues of the estimator, and $Q$ is orthonormal since the estimator is real and symmetric. Although spectral decomposition may be computationally demanding for high-dimensional problems, this operation is performed at the central server and allows to easily compute the inverse of the approximate Hessian as
\begin{equation}
    Z_k = Q \bar{D} Q^T,
    \;
    \bar{D}_{ii} = 1 / \max \left(\lambda_{\min}, \min(D_{ii}, \lambda_{\max}) \right) . 
    \label{eq:H_eigenvalue_clipping}
\end{equation}
The above formula projects the eigenvalues in the interval $[\lambda_{\min}, \lambda_{\max}]$ before the inversion, where $0 < \lambda_{\min} < \lambda_{\max}$ are design parameters. 
Finally, the target variable is updated according to the approximate Newton step
\begin{equation}
    x_{k+1} = x_k - \alpha Z_k g_k.
\end{equation}
The learning rate $\alpha$ can be either a constant value or follow an increasing schedule. The latter option is preferable to promote algorithmic stability and prevent oscillations, as at the beginning the norm of the gradient is usually large and the Hessian approximation may not be sufficiently accurate.
After a few damped iterations it is desirable to bring up the stepsize to $\alpha=1$, which is the optimal value for the exact Newton method. This is justified by the fact that once the decision vector $x_k$ begins to settle the Hessian becomes almost constant. As a consequence, in virtue of \eqref{eq:Leventhal_convergence_rate} the Hessian estimator converges linearly to the true Hessian, and one recovers an almost perfect Newton step.
The pseudocode of the algorithm summarizes the main steps.

\begin{algorithm}
\caption{FedZeN}\label{alg:FedZeN}
\begin{algorithmic}
\Initialize{
     Central server (CS): Choose $x_1 \in \mathbb{R}^d$, $r > 0$, $H_1^0 \in \mathbb{R}^{d \times d}$ symmetric. \\
     Clients: Choose $\mu \geq 0$.
     }
\State % for spacing
\State CS: Broadcast $r$ and a random seed for the PRNGs.
\For{each iteration $k = 1, 2, \dots$}
    \State All nodes: Generate $[u_1 \dots u_r]$ using \eqref{eq:Stiefel}.
    \State CS: Broadcast $x_k$.
    \For{each client $i \in [n]$}
        \State Compute $c_{ij}, b_{ij} \, j \in [r]$ using (\ref{eq:c_ij},\ \ref{eq:b_ij}).
    \EndFor
    \State CS: If $k>1$ set $H_k^0 = H_{k-1}^r$.
    \State CS: Compute $g_k$, $H_k^r$ using (\ref{eq:gradient_estimator_federated},\ \ref{eq:Leventhal_federated}).
    \State CS: Compute $Z_k$ using either \eqref{eq:H_rhoIdentity} or \eqref{eq:H_eigenvalue_clipping}.
    \State CS: $x_{k+1} = x_k - \alpha Z_k g_k$.
\EndFor
\end{algorithmic}
\end{algorithm}

We now emphasize some of the strengths of FedZeN.

\textit{(i)} The core of the algorithm is the distributed estimation of the full Hessian matrix of the global objective function, which is used for preconditioning. By considering also the off-diagonal entries of the Hessian, the algorithm can preserve fast convergence also in case of highly skewed objectives. The estimation of both gradient and Hessian only requires to evaluate the local functions at an arbitrary number of points, making the algorithm suited for black-box optimization problems in which the exact derivatives are not available.

\textit{(ii)} The algorithm is conceptually simple and straightforward to be implemented, as it is self-contained and differently from \cite{Hajinezhad_Hong_Garcia_2019} does not involve solving auxiliary subproblems. Moreover, it requires a small amount of parameter tuning, as the only design parameters are $r$, $\alpha$ and the ones required for robust matrix inversion, i.e. either $\rho$ or the pair $\lambda_{\min}, \lambda_{\max}$. In particular, $r$ determines both the number of function queries and the number of scalars transmitted by each client, allowing to adapt the computational and communication cost of the algorithm to the available resources. 

\textit{(iii)} Differently from other works which require statistical similarity between the local functions, here we do not make any assumption about the relationship between the data distributions of the clients. Indeed, by estimating the derivatives of the global objective function, FedZeN naturally handles data heterogeneity between clients. Moreover, since it does not perform multiple local iterations, it does not suffer from client drift.

\textit{(iv)} Since $g_k$ and $H_k^r$ cannot be obtained without knowing the vectors $\{u_j\}$, which in turn require knowing the seed of the PRNG, our method offers an additional level of privacy provided that the initial seed is transmitted on a secure channel. In this case, a possible eavesdropper on the communication channel between the participants and the central server would not be able to obtain the estimated derivatives, as the coefficients $c_{ij}$ and $b_{ij}$ are useless by themselves. This is very important for data security, since in some cases it is possible to reconstruct raw data samples from shared gradients \cite{geiping2020inverting}.

% The inverse of the Hessian can be efficiently (if r << d) computed using Sherman-Morrison (keeping and updating also the original matrix). However here we are also using regularization or eigenvalue clipping.

\section{Convergence analysis}
In this section we derive theoretical guarantees on the performance of the proposed algorithm. Our analysis is inspired by the one in \cite{Bollapragada_Byrd_Nocedal_2019}, which deals with subsampled Newton methods.

\begin{remark}
    Below we derive rates of convergence up to zeroth-order precision, which is the smallest theoretically achievable accuracy. Once the zeroth-order estimation error becomes dominant, one can either reduce the finite-difference granularity $\mu$ or terminate the algorithm. We recall that the design parameter $\mu$ can be chosen arbitrarily small according to the available hardware, and for $\mu \rightarrow 0$ we have exact convergence.
\end{remark}

Our first result, Theorem \ref{theorem:global_linear_convergence}, concerns the improvement of the function value and shows that the algorithm enjoys global linear convergence up to zeroth-order precision.
\begin{theorem}[Global linear convergence]
     Let $Z_k$ be computed using the eigenvalue clipping formula \eqref{eq:H_eigenvalue_clipping}, let $\alpha \leq 2 \lambda_{\min} / L_1$ and define $\gamma = \frac{\alpha 2 m}{\lambda_{\max}} \left( 1 - \frac{L_1 \alpha}{2 \lambda_{\min}} \right)$. Then each iteration of FedZeN satisfies with probability $1$, i.e. for each realization of the derivative estimators, the bound
    \begin{equation*}
         f(x_{k+1}) - f(x^\star) \leq \left( 1 - \gamma \right) (f(x_k) - f(x^\star))
        + O(\mu^2) .
    \end{equation*}
    The stepsize that maximizes $\gamma$ is $\alpha^\star = \lambda_{\min} / L_1$, for which $\gamma(\alpha^\star) = (m \lambda_{\min}) / (L_1 \lambda_{\max})$.
    \label{theorem:global_linear_convergence}
\end{theorem}

\begin{proof}
    We list the main steps: $\circled{1}{0.8}$ Taylor's expansion for functions with bounded Hessian, $\circled{2}{0.8}$ add and subtract $\nabla f(x_k)$, $\circled{3}{0.8}$ Cauchy-Schwarz, $\circled{4}{0.8}$ Lemma \ref{lemma:gradient_estimator_error}, the bound $\nabla f(x) \leq L_0$ $\forall x \in \mathbb{R}^d$ provided by Assumption \ref{assumption:Lipschitz}, and the fact that by construction $1/ \lambda_{\max} \leq \norm{Z_k} \leq 1/ \lambda_{\min}$, $\circled{5}{0.8}$ the assumption on $\alpha$. 
    \begin{equation*}
        \begin{aligned}
            & f(x_{k+1}) \overset{\circled{1}{0.6}}{\leq} f(x_k) + \nabla f(x_k)^T (-\alpha Z_k g_k) + \frac{L_1}{2} \norm{\alpha Z_k g_k}^2 \\
            % row 2
            & \overset{\circled{2}{0.6}}{=} f(x_k) - \alpha \nabla f(x_k)^T Z_k (g_k - \nabla f(x_k) +\nabla f(x_k))\\
            & \quad + \frac{L_1 \alpha^2}{2} \norm { Z_k (g_k - \nabla f(x_k) + \nabla f(x_k)) }^2 \\
            % row 3
            & \overset{\circled{3}{0.6}}{\leq} f(x_k) - \alpha \nabla f(x_k)^T \left( Z_k - \frac{L_1 \alpha}{2} Z_k^2 \right) \nabla f(x_k) \\
            & \quad + \alpha \norm{ \nabla f(x_k)} \norm{ Z_k } \norm{ \nabla f(x_k) - g_k } \\
            & \quad + \frac{L_1 \alpha^2}{2} \norm{Z_k^2} \norm{g_k - \nabla f(x_k)} \left( \norm{g_k - \nabla f(x_k)} + 2 \norm{\nabla f(x_k)} \right) \\
            % row 4
            & \overset{\circled{4}{0.6}}{\leq} f(x_k) - \alpha \nabla f(x_k)^T Z_k^{1/2} \left( I - \frac{L_1 \alpha}{2} Z_k \right) Z_k^{1/2} \nabla f(x_k) \\
            & \quad + \frac{\alpha L_0}{\lambda_{\min}} \frac{d L_2 \mu^2}{6}
            + \frac{L_1 \alpha^2}{2 \lambda_{\min}^2} \frac{d L_2 \mu^2}{6} \left( \frac{d L_2 \mu^2}{6} + 2 L_0 \right) \\
            % row 5
            & \overset{\circled{5}{0.6}}{\leq} f(x_k) - \alpha \norm{ \nabla f(x_k) }^2 \left( 1 - \frac{L_1 \alpha}{2 \lambda_{\min}} \right)  \frac{1}{\lambda_{\max}}
            +
            O(\mu^2) .
        \end{aligned}
    \end{equation*}
    Recalling that $m$-strong convexity implies $\norm{\nabla f(x)}^2 \geq 2m (f(x) - f(x^\star))$ and subtracting $f(x^\star)$ from both sides the proof is concluded.
\end{proof}

As usually done for Newton-type methods, we inspect the behaviour of the algorithm in a neighborhood of the optimal solution to derive a faster convergence rate.

\begin{theorem} [Linear-quadratic local bound]
    Consider the generic $k$-th iteration of FedZeN and choose the stepsize $\alpha=1$.      
    Compute $Z_k$ using eigenvalue clipping with $\lambda_{\min}$, $\lambda_{\max}$ such that $Z_k^{-1} = H_k^r$. Then the improvement towards the global minimum $x^\star$ satisfies
    \begin{equation*}
        \begin{aligned}
            \norm{x_{k+1} - x^\star}
            & \leq \frac{L_2}{2 \lambda_{\min}} \norm{ x_{k} - x^\star }^2 
            + \frac{d L_2 \mu^2}{6 \lambda_{\min}} \\
            & \quad + \frac{ \norm{ \nabla^2 f(x_k) - H_k^r }}{\lambda_{\min}} \norm{ x_{k} - x^\star }.
        \end{aligned}
    \end{equation*}
    \label{theorem:local_convergence}
\end{theorem}

\begin{proof}
    We first isolate the contribution due to the approximation error of the derivative estimators.
    \begin{equation*}
        \begin{aligned}
            \lVert & x_{k+1} - x^\star \rVert
            = \norm{x_{k} - x^\star - Z_k g_k} \\ 
            &= \norm{ Z_k \left( Z_k^{-1}(x_{k} - x^\star) - g_k + \nabla f(x_k) - \nabla f(x_k) \right)} \\ 
            & \leq \frac{1}{\lambda_{\min}} \left[ \norm{ Z_k^{-1}(x_{k} - x^\star) - \nabla f(x_k) } + \norm{ \nabla f(x_k) - g_k} \right] \\
            & \leq \frac{1}{\lambda_{\min}} \left[ \norm{ (H_k^r - \nabla^2 f(x_k)) (x_{k} - x^\star) } + \norm{ \nabla f(x_k) - g_k} \right]\\
            & \quad + \frac{1}{\lambda_{\min}} \norm{ \nabla^2 f(x_k) (x_{k} - x^\star) - \nabla f(x_k)}.
        \end{aligned}
    \end{equation*}
    Recalling that $\nabla f(x^\star)=0$ and using the fundamental theorem of calculus and the Lipschitz property of the Hessian, we can bound the norm in the last term as
    % see "Nocedal Wright - Numerical Optimization" Theorem 3.5
    \begin{equation*}
        \begin{aligned}
            \lVert & \nabla^2 f(x_k) (x_{k} - x^\star) + \nabla f(x^\star) - \nabla f(x_k) \rVert \\
            & \leq \norm{ \nabla^2 f(x_k) (x_k - x^\star) + \int_0^1 \frac{d}{dt} \nabla f(x_k + t(x^\star - x_k)) \; dt } \\
            &= \norm{ \int_0^1 \left[ \nabla^2 f(x_k) - \nabla^2 f(x_k + t(x^\star - x_k))  \right] ( x_k - x^\star ) \; dt} \\
            & \leq \norm{ x_k - x^\star } \int_0^1 \norm{ \nabla^2 f(x_k) -\nabla^2 f(x_k + t(x^\star - x_k)) } \; dt \\
            & \leq \norm{ x_{k} - x^\star } \int_0^1 L_2 \norm{ t (x_{k} - x^\star) } \; dt \\
            &= \frac{L_2 \norm{ x_{k} - x^\star }^2 }{2}.
        \end{aligned}
    \end{equation*}
    Using the above result and Lemma \ref{lemma:gradient_estimator_error} we get the bound to be proved.
\end{proof}

The bound provided by Theorem \ref{theorem:local_convergence} can be used to prove local quadratic convergence up to zeroth-order precision with high probability. To do so, from now on we allow a variable number of search directions at each iteration. To simplify the analysis, we bound with high probability the approximation error of the Hessian estimator by means of the following assumption.

\begin{assumption}
    Fix a symmetric $H^0 \in \mathbb{R}^{d \times d}$, an accuracy $\epsilon_k > 0$ and a failure probability $\delta \in (0,1)$. Then there exist a finite-difference precision $\bar{\mu}(\epsilon_k, \delta)$ and a minimum number of search directions $\bar{r}(\epsilon_k, \delta, \mu)$ such that $\forall \mu \leq \bar{\mu}(\epsilon_k, \delta)$ and $\forall r \geq \bar{r}(\epsilon_k, \delta, \mu)$ the Hessian estimator $H^r$ defined in \eqref{eq:Leventhal} satisfies $\mathbb{P} \left( \norm{\nabla^2 f(x_k) - H^r} \geq \epsilon_k \right) \leq \delta$.
    \label{assumption:Hessian_estimator_error}
\end{assumption}

\begin{remark}
    In the interest of space, we do not derive an explicit formula for $\bar{\mu}(\epsilon_k, \delta)$ and $\bar{r}(\epsilon_k, \delta, \mu)$. Rigorous bounds on the minimum number of search directions needed to satisfy the inequality in Assumption \ref{assumption:Hessian_estimator_error} for any value of $(\epsilon, \delta, \mu, H^0)$ will be provided in future work. Intuitively, by shrinking $\mu$ one can make the zeroth-order error negligible and approximate arbitrarily well the update \eqref{eq:Leventhal}. The latter is known to converge almost surely to the true Hessian, and the existence of $\bar{r}$ is guaranteed by the convergence rate \eqref{eq:Leventhal_convergence_rate}.
\end{remark}

\begin{theorem} [Local quadratic convergence w.h.p.]
    Consider the assumptions of Theorem \ref{theorem:local_convergence} and let Assumption \ref{assumption:Hessian_estimator_error} be satisfied. At each iteration $k$ of FedZeN:
    
    \begin{itemize}
        \item If $\norm{g_k} > \frac{d L_2 \mu^2}{6}$, according to Assumption \ref{assumption:Hessian_estimator_error} choose $\delta \in (0,1)$, $\epsilon_k \leq \frac{1}{L_1} \left( \norm{g_k} - \frac{d L_2 \mu^2}{6} \right)$, and let $\bar{r}(\epsilon_k, \delta, \mu)$ be the corresponding minimum number of search directions. If $r > \bar{r}(\epsilon_k, \delta, \mu)$, then with probability $(1 - \delta)$ the local convergence rate up to zeroth-order precision is quadratic:
        \begin{equation*}
                \begin{aligned}
                    \norm{x_{k+1} - x^\star}
                    & \leq \frac{L_2 + 2}{2 \lambda_{\min}} \norm{ x_{k} - x^\star }^2 
                    + O(\mu^2).
                \end{aligned}
            \end{equation*}
    
        \item If $\norm{g_k} \leq \frac{d L_2 \mu^2}{6}$, then the suboptimality gap satisfies
        \begin{equation*}
            \norm{x_k - x^\star}
            \leq \frac{d L_2 \mu^2}{3 m} = O(\mu^2)
        \end{equation*}
        and one can use this as stopping criterion.
    \end{itemize}
    \label{theorem:quadratic}
\end{theorem}

\begin{proof}
    In the first case, Assumption \ref{assumption:Hessian_estimator_error} guarantees that with probability $(1 - \delta)$ it holds
    \begin{equation*}
        \begin{aligned}
            & \norm{\nabla^2 f(x_k) - H^r}
            \leq \epsilon_k
            \leq \frac{ \norm{g_k} - \frac{d L_2 \mu^2}{6} }{L_1} \\
            & \quad \leq \frac{ \left| \norm{g_k} - \norm{\nabla f(x_k) - g_k} \right| }{L_1}
            \leq \frac{\norm{\nabla f(x_k)}}{L_1}
            \leq \norm{ x_{k} - x^\star } .
        \end{aligned}
    \end{equation*}
    Combining the above inequality with Theorem \ref{theorem:local_convergence} we obtain the quadratic rate to be proved.
    In the second case, we have
    \begin{equation*}
        \begin{aligned}
            & \norm{ x_{k} - x^\star }
            \leq \frac{\norm{\nabla f(x_k)}}{m}
            \leq \frac{\norm{g_k} + \norm{\nabla f(x_k) - g_k}}{m} \\
            & \quad \leq \frac{\norm{g_k} + \frac{d L_2 \mu^2}{6}}{m}
            \leq \frac{2}{m} \frac{d L_2 \mu^2}{6}.
        \end{aligned}
    \end{equation*}
\end{proof}

\begin{remark}
    The condition on $\epsilon_k$ used in Theorem \ref{theorem:quadratic} is implementable in practice by following these steps: \textit{(i)} search along $d$ orthonormal directions and build $g_k$, \textit{(ii)} use the latter to compute the upper bound on $\epsilon_k$, \textit{(iii)} choose $r \geq \bar{r}(\epsilon_k, \delta, \mu)$, \textit{(iv)} evaluate the function along the remaining $r - d$ directions to build $H_k^r$.
\end{remark}

% Note: There is the issue of the high probability, at some iterations we may lose the property. It is not truly quadratic. However, 1) we still have global linear convergence at all iterations, 2) it is a local bound, which implies that we are close to the optimum and the Hessian to estimate is almost constant.

\section{Numerical results}
We start the numerical analysis by comparing the proposed distributed zeroth-order Hessian estimator to the main Hessian estimators available in the literature. We consider the following competitors: \textit{(i)} the identity matrix, which is implicitly used by the methods that only use gradient estimates, \textit{(ii)} the Jacobi estimator employed in \cite{Maritan_Schenato_2023}, which approximates the diagonal of the Hessian matrix, \textit{(iii)} a distributed version of the randomized estimator based on the second-order Stein's identity proposed in \cite{Balasubramanian_Ghadimi_2022}, and \textit{(iv)} a distributed version of the one based on Stiefel sampling, introduced by \cite{Feng_Wang_2023}. Differently from our Hessian estimator, all the aforementioned ones are thought to be reset at each iteration. To allow a complete comparison, we also implement incremental versions of the last two estimators where the latest estimate is used as starting point.

\begin{figure}[h!]
    \centering
 
    \begin{subfigure}[b]{\columnwidth}
        \centering
        \parbox{0.9\columnwidth}{
             \includegraphics[width=0.95\columnwidth]{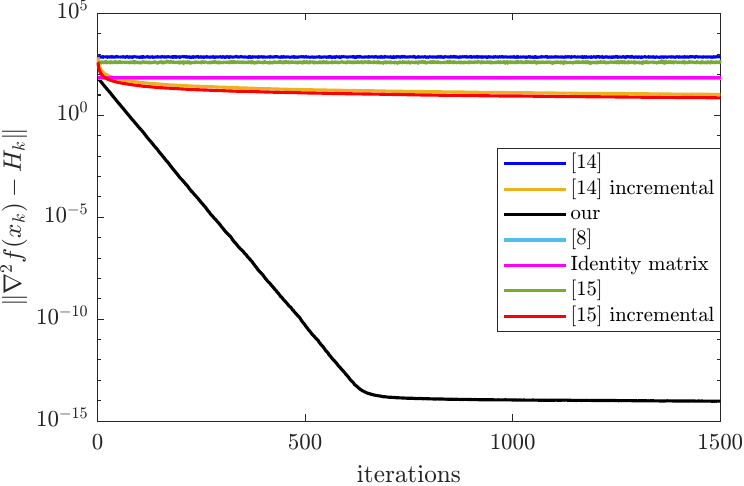}
        }
    \end{subfigure}

    \caption{Approximation error of various Hessian estimators in case of constant Hessian, averaging over $100$ matrices. 
    }
    \label{fig:constant_H_error}
  
\end{figure}

Figure \ref{fig:constant_H_error} displays the evolution of the approximation error in case of constant Hessian, showing that our distributed zeroth-order version of \eqref{eq:Leventhal} outperforms all the other estimators, including the incremental versions of the competitors. This happens because while the estimators \cite{Feng_Wang_2023} and \cite{Balasubramanian_Ghadimi_2022} are sample averages, the update \eqref{eq:Leventhal} imposes the correct curvature along each search direction.
The plot shows the average errors over $100$ random Hessian matrices. All algorithms perform the same number of function evaluations per iteration, namely $2d+1$, which is the amount of queries required by the deterministic Jacobi estimator \cite{Maritan_Schenato_2023}.
Since the Hessian is known to be constant, the incremental versions of \cite{Feng_Wang_2023} and \cite{Balasubramanian_Ghadimi_2022} compute the mean of all the past estimates, e.g. $H_k^{\text{inc}} = (H_k + (k-1) H_{k-1})/k$.

\begin{figure}[h!]
    \centering
 
    \begin{subfigure}[b]{0.9\columnwidth}
        \centering
        \parbox{\columnwidth}{
        \hspace{1cm}
        \includegraphics[width=0.9\columnwidth]{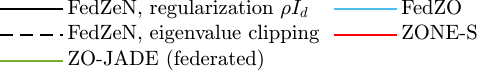}
        }
    \end{subfigure}
     
    \vspace{0.2cm}
    \hfill
    \begin{subfigure}[b]{\columnwidth}
        \centering
        \parbox{0.9\columnwidth}{\includegraphics[width=0.95\columnwidth]{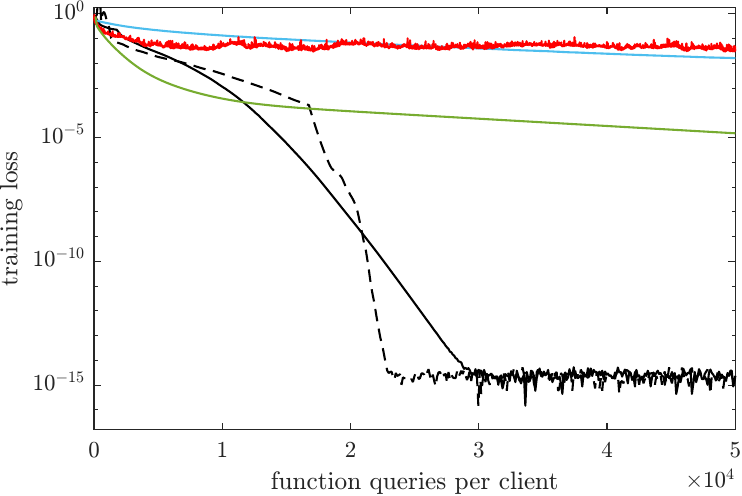}
        }
    \end{subfigure}

    \vspace{0.2cm}
    \hfill
    \begin{subfigure}[b]{\columnwidth}
        \centering
        \parbox{0.9\columnwidth}{\includegraphics[width=0.95\columnwidth]{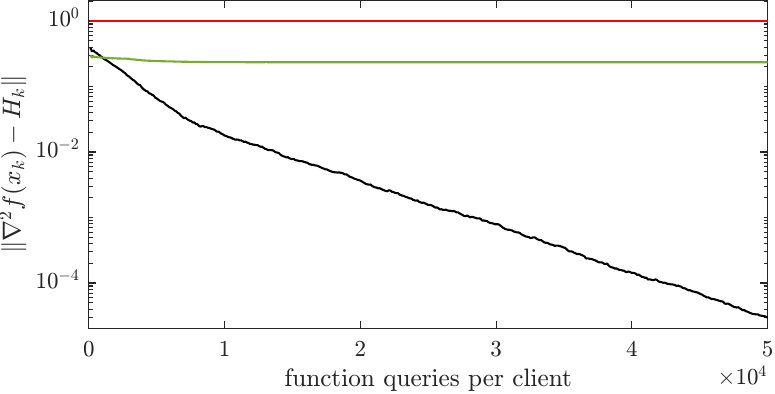}
        }
    \end{subfigure}

    \caption{Logistic regression using the dataset Covertype \cite{misc_covertype_31}. FedZeN (our) versus other federated zeroth-order algorithms.}
    \label{fig:vs_ZO}
  
\end{figure}

We now empirically test the efficiency of the proposed FedZeN. We compare it to the zeroth-order federated algorithms FedZO \cite{Fang_Yu_Jiang_Shi_Jones_Zhou_2022} and ZONE-S \cite{Hajinezhad_Hong_Garcia_2019} and to a federated version of ZO-JADE \cite{Maritan_Schenato_2023}, which is designed for general mesh networks of agents.
We evaluate the performances of FedZeN setting $\lambda_{\min} = 10^{-3}$, $\lambda_{\max} = 10^{4}$, $\rho = 10^{-2}$. For the competitor algorithms we try several hyperparameter configurations to find the one that leads to the best performance.
In Figure \ref{fig:vs_ZO} we show FedZO with learning rate $\eta=0.1$ and $H=10$ local epochs. Regarding ZONE-S, the augmented Lagrangian is minimized using Nesterov accelerated gradient, and since only one client is active at each iteration we show the average number of function queries. All algorithms are started from the same initial $x_1$ and use $r=d=55$ search directions.

The test is a binary classification problem via logistic regression, where the dataset Covertype \cite{misc_covertype_31} is evenly split over a pool of $n=10$ clients. The local objectives are the regularized log-losses
\begin{equation*}
f_i(x, \mathcal{D}_i) = \frac{1}{\mathcal{D}_i} \sum_{k=1}^{|\mathcal{D}_i|} \log\left( 1 + \exp\left( - l_k [s_k^T \ 1] x \right) \right) + \frac{w}{2} \left\| x \right\|^2,
\end{equation*}
where $l_k \in \{-1,1\}$ is the label associated to the sample $s_k \in \mathbb{R}^{d-1}$, $d=55$ and $w > 0$. The normalized training loss shown in the plots is $ \left( f(x) - f(x^\star) \right) / |f(x^\star)| $, where $x^\star$ is the global minimum.

%+++++

Figure \ref{fig:vs_ZO} clearly shows that FedZeN outperforms the other zeroth-order algorithms, which instead settle far from the optimal solution. This was to be expected, as methods that use only the gradient typically converge much more slowly than those that also exploit the Hessian. The speed of convergence is expressed in terms of number of function evaluations, which in the ZO optimization field are assumed to be the most expensive computations. Plotting the training loss against the number of iterations or the the number of scalars transmitted and received by each client, one obtains figures identical to the one shown. The plot on the bottom confirms the effectiveness of the incremental Hessian estimator employed in FedZeN when the target Hessian changes over time. Since methods that estimate only the gradient can be thought to use the identity matrix as Hessian estimator, in ZONE-S and FedZO we set $H_k = I_d$ $\forall k$ just for reference.

% Figure \ref{fig:vs_nonZO} shows that while being a zeroth-order algorithm, FedZeN can compete even with second-order methods that have access to the exact derivatives. Moreover, while randomized Hessian estimators usually require $O(d^2)$ function evaluations to provide a good approximation, here we already obtain good performances by setting $r=3d$ with $d=55$.

\section{Conclusions}
We have introduced a general procedure to estimate the global Hessian matrix in the federated learning setting when exact derivatives are not available and functions are only accessible through point evaluations. Under the mild assumption that all nodes own a pseudo-random number generator, we generate a common set of search directions at all the nodes, sampling from the Stiefel manifold for greater estimation accuracy. This allows to greatly reduce the communication complexity and conceal the estimated derivatives from external eavesdroppers. Since the Hessian estimator is incremental and builds upon past estimates, few function evaluations per iteration are required. This distributed estimation technique is the foundation of the proposed FedZeN, a zeroth-order algorithm for federated learning which is the first to approximate and leverage the Hessian matrix. FedZeN allows to tailor both the communication and the computational costs to the capabilities of the clients by selecting an appropriate number of search directions. Moreover, the algorithm is suited for federations of clients with heterogeneous data distributions. FedZeN comes with theoretical guarantees of global linear and local quadratic convergence up to zeroth-order precision. Numerical simulations confirm that FedZeN converges superlinearly, outperforming the main federated zeroth-order algorithms.

\bibliographystyle{IEEEtran}
\bibliography{references}

\end{document}